%% file: main.tex
\documentclass[10pt,journal,compsoc]{IEEEtran}
\usepackage{ifthen}
\usepackage{svg}
\usepackage{svg-extract}
\usepackage{efbox,graphicx}
\usepackage{url}
\usepackage{multirow}
\usepackage{array} 
\usepackage{amsmath}
\usepackage{amssymb}
\usepackage{graphicx}
\usepackage{flushend}
\usepackage{enumitem}
\usepackage{xcolor}
\usepackage{svg}
\usepackage{svg-extract}
\usepackage{subcaption}
\usepackage{xspace}
\usepackage{tabulary}
\usepackage{booktabs}
\usepackage{efbox,graphicx}
\usepackage{amsmath,amssymb,latexsym}
\usepackage{csvsimple}
\usepackage{longtable}
\usepackage{array,longtable}
\usepackage{multirow}
\usepackage{float}
\usepackage{enumitem}
\usepackage{stfloats}
\usepackage{url}
\usepackage{graphicx}
\usepackage{caption}
\usepackage{subcaption}
\usepackage{amssymb}
\usepackage{amsmath}
\usepackage{amsthm}
\usepackage{makecell}
\usepackage{outlines}
\usepackage{array,ragged2e}
\newcolumntype{P}[1]{>{\raggedright \arraybackslash}p{#1}}

\ifCLASSOPTIONcompsoc
  \usepackage[nocompress]{cite}
\else
  \usepackage{cite}
\fi

\input{memosetup}

\begin{document}
\title{Leveraging Human Salience to Improve Calorie Estimation}

\author{
Katherine Dearstyne, Alberto Rodriguez
 \\
\IEEEmembership{Department of Computer Science and Engineering, University of Notre Dame} \\
\IEEEmembership{ Notre Dame, Indiana 46556 USA} \\
\IEEEmembership{\{kdearsty, arodri39\}@nd.edu}
}

\IEEEtitleabstractindextext{%
\begin{abstract}
\input{chapters/abstract}
\end{abstract}

}

\maketitle

\section{Introduction}\label{sec:intro}
\input{chapters/intro}

\section{Experiment Design}\label{sec:design}
\input{chapters/design}

\section{CYBORG: Integrating Human Saliency Into Calorie Estimation}\label{sec:cyborg}
\input{chapters/cyborg}

\section{Results}\label{sec:results}
\input{chapters/results}

\section{Conclusion}\label{sec:conclusion}
\input{chapters/conclusion.tex}

\bibliographystyle{IEEEtran}
\bibliography{bibliography} 

\end{document}

%% file: memosetup.tex
\newboolean{showcomments}
\setboolean{showcomments}{true} % toggle to show or hide comments
\ifthenelse{\boolean{showcomments}}
  {

  }

%% file: chapters/abstract.tex
The following paper investigates the effectiveness of incorporating human salience into the task of calorie prediction from images of food. We observe a 32.2\% relative improvement when incorporating saliency maps on the images of food highlighting the most calorie regions. We also attempt to further improve the accuracy by starting the best models using pre-trained weights on similar tasks of mass estimation and food classification. However, we observe no improvement. Surprisingly, we also find that our best model was not able to surpass the original performance published alongside the test dataset, Nutrition5k \cite{thames2021nutrition5k}. We use ResNet50 and Xception as the base models for our experiment.

%% file: chapters/intro.tex
As obesity and other food-related illnesses affect a growing proportion of society, there is an increasing need for individuals to become more aware of what foods they consume. The science of nutrition is complex and often times inconclusive, yet the simple formula for managing weight and health can be boiled down to: calories in = calories out. Since calculating calories for home cooked meals is often tedious and there is generally little nutrition information available for food ordered from restaurants, monitoring calorie consumption can be a difficult task for most people. For these reasons, there is a need to automatically capture calorie information with as little input from a user as possible. Therefore, we propose a model that can predict the number of calories in a dish with just a single image of the plate of food. However, this is a difficult challenge for many reasons including: 

\begin{itemize}
    \item Calorie calculations are not only dependent on the type of food but also the volume of the food, which is often hard to detect within a single image.
    \item Food is taken across many different cameras resulting in differences in quality, angles, and lighting.
    \item There are presently very few datasets containing images of food items and their calorie counts.
\end{itemize}

In order to overcome these challenges, a model would have to learn to extract features that are agnostic to specific food types, background scenery, and quantity. Previously, we investigated using several pre-training steps, namely food classification and mass estimation, to enhance the model's ability to predict calories. We build off this work by investigating whether using human saliency maps (HSMs) (shown in Figure \ref{fig:example}) in our training process can guide the model towards features human deem most important. This is inspired by the success of the CYBORG loss function \cite{cyborg} when distinguishing artifically generated images of faces from real ones. For the goal of calorie prediction, we adapt this framework to incorporate human judgements of image regions containing foods with the highest number of calories.

Therefore, we present several experiments that seek to address our four primary research questions:

\begin{enumerate}[label={RQ\arabic*:}]
    \item Do HSMs improve the performance of calorie estimation?
    \item Can pre-training tasks improve the performance of these models that use HSMs?
    \item Can an ensemble model composed of the best models improve performance?
    \item How does our best model compare against the performance of the model published on Nutrition5k (test set)?
\end{enumerate}

\input{figs/calorie_cam.tex}

%% file: figs/calorie_cam.tex
\begin{figure}
\centering
\begin{subfigure}{.5\columnwidth}
  \centering
  \includegraphics[width=.9\columnwidth]{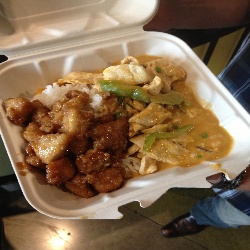}
  \label{fig:source_example}
\end{subfigure}%
\begin{subfigure}{.5\columnwidth}
  \centering
  \includegraphics[width=.9\columnwidth]{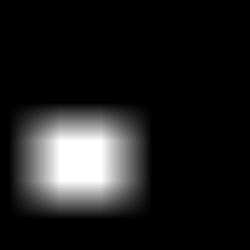}
  \label{fig:hmap_example}
\end{subfigure}
\caption{Example of an image used for calorie estimation alongside the human saliency map indicating the most caloric regions of the food.}
\label{fig:example}
\end{figure}

%% file: chapters/design.tex
To address our research questions, we perform four different experiments: 

\begin{enumerate}[label=\arabic*.]
    \item To answer RQ1, we evaluate the effectiveness of the HSMs (described in Section \ref{sec:cyborg}) by comparing two model architectures (see Section \ref{sec:models}) on the calorie estimation task both with and without the HSMs.
    \item To answer RQ2, we evaluate if the tasks of food categorization and mass estimation provide better starting weights than ImageNet for training on the calorie estimation task with integrated HSMs. Each model architecture is first trained on the food categorization or mass estimation task, and then the weights of the best-performing models are used as the initial weights for the calorie prediction task.
    \item To answer RQ3, we take the best two models and combine them in an ensemble. Prefacing our results, this turned out to be two Xception models, with one trained for mass prediction and the other for calorie prediction but leveraging HSMs.
    \item To answer RQ4, we take the best model from all of the research questions and fine-tune it on Nutrition5k. Up to this research question, our experiments have been using Nutrition5k as a hold-out test set. However, the original paper \cite{thames2021nutrition5k} reports their performance after training on 80\% of the dataset and testing on 20\%. They have published their train and test splits, thus, we take our best model and attempt to replicate their work to compare the results. Namely, for this research question we compare two models on the train split, one starting from ImageNet weights and the other resulting from the best performance from our prior experiments.
\end{enumerate}
Below, we outline the data, model architectures, and metrics used in these experiments.
\subsection{Data}
\label{sec:data}
In total, we use four different datasets containing images of food which are described in Table \ref{tab:datasets_overview} alongside their usage within the experiment. The main calorie prediction task is conducted with MenuMatch \cite{menu_match} as the training and validation set while Nutrition5k\cite{thames2021nutrition5k} is used as the hold out test set. FoodImages \cite{noauthor_food_nodate} is used as the training and validation sets for the task of food classification. Finally, ECUST \cite{liang2017computer} is used as the training and validation sets for the mass estimation task. All training sets consist of 80\% of the available training data and the remaining is used for validation. Note, the food classification and mass tasks don't contain test sets because they are used \textit{only} as starting weights on the calorie prediction task.
\input{tables/datasets_overview}

\subsection{Neural Models}
\label{sec:models}
\begin{figure}
    \centering
    \includegraphics[width=\columnwidth]{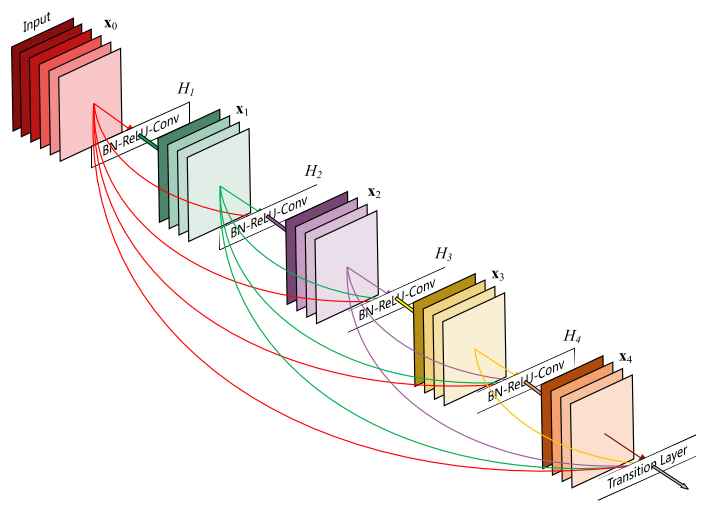}
    \caption{ResNet model architecture.}
    \label{fig:resnet_architecture}
\end{figure}

\begin{figure}
    \centering
    \includegraphics[width=\columnwidth]{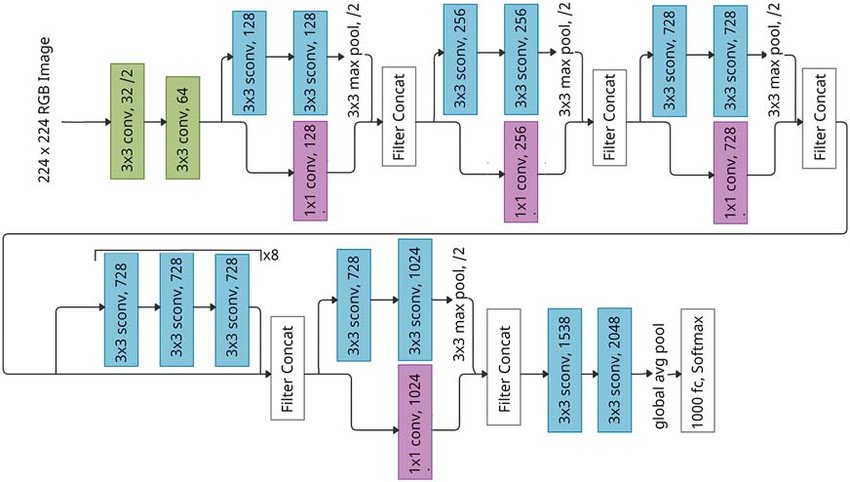}
    \caption{Xception model architecture.}
    \label{fig:xception_architecture}
\end{figure}

\paragraph{ResNET} The residual network model, otherwise known as ResNet, is deep convolutional model based from the VGG architecture but which leverages skip connects to jump over some layers -- mimicking the biology of our brain. Figure \ref{fig:resnet_architecture} displays a more details view of the underlying architecture of the model.
\paragraph{Xception} Xception is a convolution neural model based on depthwise separable convolution layers \cite{Chollet_2017}. Figure \ref{fig:xception_architecture} described in the detail the layers associated in this model. This model relies on the assumption that the mapping of cross-channels correlations and spatial correlations in the feature maps of convolutional neural networks can be entirely decoupled. Xception is an more robust and efficient version of the Inception model.
We plan adding a final dense layer to the models to match the number of food categories spanning those in our pre-training task. Then, for the fine-tuning task we plan on replacing this layer with a single node predicting the number of calories present in the image. We plan on using the Keras implementation of the VGG16 and ResNet models respectively \cite{vgg_keras, resnet_keras}.
\paragraph{Ensemble} For creating an ensemble of two or more models, using the Keras graph API we take the output of all the models and concatenate them into a tensor. Then, the tensor is passed into a 100 neuron dense hidden layer before feeding into the last dense layer composed of a single neuron for regression.

\subsection{Metrics}
The first pre-training task focuses on predicting what categories of food are present in an image. Our predictions will be represented by an encoded vector of size $n$ containing 1 wherever a food category is present and 0 otherwise where $n$ is the number of food classes spanning our pre-training data. We plan to use the \textbf{categorical crossentropy} loss function along with \textbf{log loss} as our training metrics based on common uses in the field \cite{mishra_metrics_2020, t_comprehensive_2021}.
Meanwhile, the second pre-training task (predicting volume) as well as our primary task (calorie estimation) are both regression problems. Therefore, we are choosing to use the \textbf{Root Mean Squared Error} as our training metric. We use the \textbf{mean squared error} loss function to train the mass prediction task as recommended by other practitioners \cite{brownlee_regression_2021}. However, we compare the results of using the \textbf{mean squared error} loss function alone with the results of using the CYBORG loss function \cite{cyborg} for the calorie classification.
\begin{figure*}
    \centering
    \includegraphics[width=.9\textwidth]{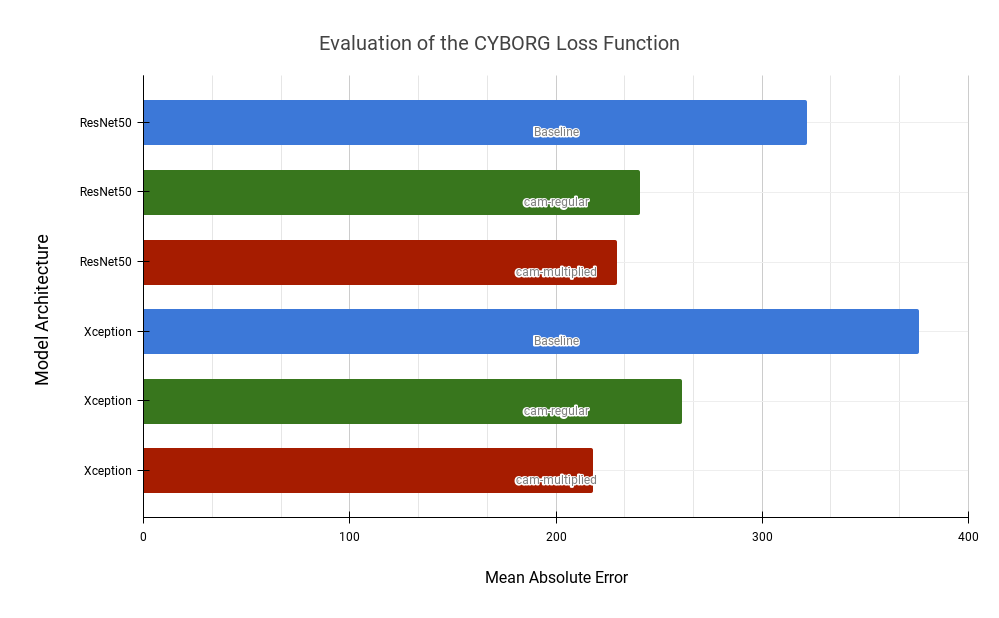}
    \caption{Evaluation metrics for the CYBORG loss function (RQ1).}
    \label{fig:rq1_results}
\end{figure*}

%% file: tables/datasets_overview.tex
\begin{table*}
    \centering
    \bgroup
    \def\arraystretch{1.5}
        \begin{tabular}{|P{3cm}|P{4cm}|P{1cm}|P{3cm}|}
            \hline
            \textbf{Name} & \textbf{Description} & \textbf{Size} & \textbf{Usage }\\ \hline
            Food Image Classification Data \cite{noauthor_food_nodate} & Images of food spanning over 101 categories. & 101K & Food classification train and validation set.\\ \hline
            % UNIMIB2016 \cite{cioccaJBHI} & Tray images with multiple foods and containing 15 food categories. & 2K \\ \hline
            ECUST Food Dataset \cite{liang2017computer} & Images of food, their calorie counts and volume & 3K & Mass estimation train and validation set. \\ \hline
            MenuMatch \cite{menu_match} & Images of food and their calorie counts. & 646 & Calorie estimation train and validation set. \\ \hline
            Nutrition5k \cite{thames2021nutrition5k} & Images of food, their calorie counts, ingredients and weight. & 5k & Hold-out test set for calorie estimation.\\ \hline
        \end{tabular}
    \egroup
    \caption{Datasets used during the pre-training and regular training of our neural models.}
    \label{tab:datasets_overview}
\end{table*}

%% file: chapters/cyborg.tex
This section describes the integration of HSMs into the training process for calorie estimation. Previously, in the CYBORG loss function \cite{cyborg}, saliency maps were compared against a model's class activation mappings (CAMs) to penalize the model for focusing on regions deemed less important to the task by humans. However, this loss function assumes the task to be a classification problem whereas calorie prediction is a regression task. Therefore, we adapt the CYBORG loss function for regression tasks as described below.

To first create the HSMs, we instructed workers on Amazon's Mechanical Turk to create bounding boxes around the most caloric sections in the MenuMatch dataset images \cite{menu_match}. These results are converted into heatmaps by translating the areas captured by the box bounding into high value pixels and setting the remaining areas to pixel values of 0. Due to budget constraints by the AWS team, we were prevented from running the experiment multiple times. Instead we apply blurring to the bounding box, using a large kernel size (250x250), operating on the assumption that the center region of the bounding boxes contains the most important information. 

To obtain the model's saliency mappings (MSMs), we extract the feature maps from the final convolutional layer of the model as well as their associated weights in the final dense layer which contains a single neuron for the regression output. We then calculate a weighted average of these feature maps to summarize the model’s salient regions for each sample. The HSMs are resized to match the MSMs (7x7), and then both are normalize to the range of 0 and 1, using min-max scaling. 

Finally, the loss between the MSMs and HSMs as well as the loss from the predicted and true calories are both calculated using the mean squared error (MSE). In order to combine these two losses ($L_{total}$), we explore two different methods. The first method, \emph{cyborg-original}, uses a weighted sum (see Equation \ref{eq1}), with equal weight on both the calorie and heatmap loss (alpha=0.5).
\begin{equation}\label{eq1}
    L_{total} = (1-\alpha)*L_{m} - \alpha*L_{c}
\end{equation}
where $L_{m}$ is the loss between the saliency maps, $L_{c}$ is the calorie loss and $\alpha$ is the weight factor.
Since the calorie loss can easily reach 100K while the heatmap loss remains under 1, the weighted sum heavily favored the calorie estimation without the benefit of the feature map. The second method, \emph{cyborg-multipled}, addresses this issue by multiplying both losses together as shown in equation \ref{eq2}.
\begin{equation}\label{eq2}
    L_{total} = L_{m}*L_{c}
\end{equation}

%% file: chapters/results.tex
The section below describes the results of the experiments answering RQ1-4. We note that for each of these questions the reported MAE is on the entire Nutrition5k dataset (RQ1-3) or a section of it (RQ4). 

\input{tables/rq1_results}
\input{tables/rq2_results}
\input{tables/rq3_results}
\input{tables/rq4_results}
\subsection{RQ1: Do human saliency maps improve the results of calorie
estimation?}
Figure \ref{fig:rq1_results} displays the raw results of the 6 different models with the blue bars representing the baseline methods, the green showing the models with \emph{cyborg-original} loss function, and the red highlighting the models  using \emph{cyborg-multipled} loss. Table \ref{tab:rq1_results} displays the relative improvement of each method over the best baseline architecture, Xception.
We find that the models using the CYBORG loss function significantly outperformed the baseline models with an average relative improvement of 25.3\% over the best baseline method on the unseen test set. Specifically, the greatest improvement came from the models utilizing the adjusted CYBORG function which combined calorie loss with the feature map loss by multiplying them together.

\subsection{RQ2: Can pre-training tasks improve the accuracy of these models that use human saliency maps?}
Table \ref{tab:rq2_results} shows the performance of pre-trained models with the \emph{cyborg-multipled} loss function used in the fine-tuning step. Although all the alternatives surpassed the best baseline model (i.e. no pre-training or CYBORG loss), there is little to no improvement over any of the models using the CYBORG loss in Experiment 1, and none of the models from Experiment 2 outperform the best model from 1. This signals that starting with ImageNet weights is not a bad weight initialization scheme. However, we leave for future work the exploration of why these weights did not help the model improve its predictions.

\subsection{RQ3: Can an ensemble model composed of the best models improve performance?} Table \ref{tab:rq3_results} shows the performance of an ensemble model composed of two models. First, an Xception model pre-trained on mass estimation and then on the MenuMatch dataset for calorie estimation. Second, an Xception model trained on calorie estimation on the MenuMatch dataset and leveraging the HSMs and the CYBORG-multiplied loss function. We notice that the ensemble model did not do better than the best baseline model, which is surprising considering that one of its models achieved the best performance (CYBORG-multiplied).

\subsection{RQ4: How does our best model compare against that the performance published on Nutrition5k (test set)?} Table \ref{tab:rq4_results} displays the performance of three different models after training and testing on Nutrition5k, which was previously our hold out test set. The first row, Xception (base) is an Xception model initialized with ImageNet weights and trained as described. The second row, CYBORG-multiplied is our Xception model trained on calorie estimation using the Human Saliency Maps (HSMs) and performing the best on the entire Nutrition5k dataset (zero shot). The final row is the results published alongside the original Nutrition5k paper \cite{thames2021nutrition5k}. Surprisingly, not only were we not able to perform better than the published results, but our CYBORG model was not able to surpass the baseline model trained on only Nutrition5k train set. 

%% file: tables/rq1_results.tex
\begin{table}[H]
    \centering
    \begin{tabular}{|c|c|c|c|}
    \hline
         \textbf{Method} &  \textbf{Model} & \textbf{MAE} & \textbf{Improvement} \\ \hline
         Baseline & ResNet50 & 321.60 & 0.0\% \\ \hline
         Baseline & Xception & 376.0 & -16.9\% \\ \hline \hline
         CYBORG-Regular & ResNet50 & 240.65 & 25.17\% \\ \hline
         CYBORG-Regular & Xception & 261.16 & 18.79\% \\ \hline \hline
         CYBORG-multiplied & ResNet50 & 229.67 & 28.59\% \\ \hline
         \textbf{CYBORG-multiplied} & \textbf{Xception} & \textbf{214.05} & \textbf{32.18\%} \\ \hline
    \end{tabular}
    \caption{The performance of baseline models alongside those using CYBORG loss function.}
    \label{tab:rq1_results}
\end{table}

%% file: tables/rq2_results.tex
\begin{table}[H]
    \centering
    \begin{tabular}{|c|c|c|c|}
    \hline
         \textbf{Task} &  \textbf{Model} & \textbf{MAE} & \textbf{Improvement} \\ \hline
         Mass Estimation & ResNet50 & 228.13 & 29.06\% \\ \hline
         Mass Estimation & Xception & 301.09 & 6.38\% \\ \hline \hline
         Food Classification & ResNet50 & 228.56 & 28.93\% \\ \hline
         Food Classification & Xception & 255.43 & 25.91\% \\ \hline
    \end{tabular}
    \caption{The performance of models pre-trained on analogous tasks before fine-tuned using the CYBORG loss function.}
    \label{tab:rq2_results}
\end{table}

%% file: tables/rq3_results.tex
\begin{table}[H]
    \centering
    \begin{tabular}{|c|c|c|c|}
    \hline
         \textbf{Mass Model} & \textbf{CYBORG Model} & \textbf{MAE} & \textbf{Improvement} \\ \hline
         Xception & Xception (multiplied) & 376.98 & -14.69\% \\ \hline
    \end{tabular}
    \caption{The performance of the ensemble model (composed of two models) trained on the MenuMatch dataset and tested on the Nutrition5k dataset. \emph{The ensemble model was composed of the CYBORG-multiplied model trained on calorie prediction and the Xception model trained on mass estimation then calorie estimation.}}
    \label{tab:rq3_results}
\end{table}

%% file: tables/rq4_results.tex
\begin{table}[H]
    \centering
    \begin{tabular}{|c|c|c|}
    \hline
         \textbf{Model} & \textbf{MAE}\\ \hline
         Xception (base) & 217.97 \\ \hline
         CYBORG-multiplied & 225.29 \\ \hline
         Published Results & 150.8 \\ \hline
    \end{tabular}
    \caption{The performance of a baseline model, CYBORG-multiplied model, alongside the original results after training and testing on Nutrition5k using the published sets.}
    \label{tab:rq4_results}
\end{table}

%% file: chapters/conclusion.tex
Our results show that incorporating human judgements into a calorie predictor model can improve results. When using HSMs during training, we achieved as much as a 32\% decrease in MAE on the test data. This is likely because it forced the model to focus on features which generalized well across different datasets. Nonetheless, there is still much to explore regarding how best to harness the power of these human judgements. As we saw in RQ3-4, using the starting weights of the best HSM model and then performing further training resulted in a decrease in performance over even the baseline model. One possible explanation is that once the model was no longer being penalized for focusing on other regions of the image, it began "forgetting" (or not utilizing) these features. This could be explored further by capturing what areas have the highest activation after additional training or by freezing some of the layers in order to retain what was learned during training with the HSMs.

In most of our experiments we saw a much lower MAE (around 122 calories less) for validation than we did on the test set. This is not surprising since each dataset contained very different food categories that presented a challenge for the model. More interesting, however, is that there was about the same difference in calories between the validation and test set when we were using Nutrition5k for both. What is surprising is that the model did not outperform the previous experiments where it was tested on an entirely unseen dataset with very distinct food categories. One possible explanation could be that Nutrition5k is one of the few datasets to contain images at wildly different angles. Perhaps data augmentation with random rotations on the training dataset could help improve these results. 

In conclusion, human salience shows promise for improving calorie predictions but more exploration needs to be done to establish the method of training that can result in the best generalization on real-world data. Nonetheless, with an average error of around 200 calories, the model can still help humans have a reasonable calorie estimation of their plate of food, and we are optimistic these estimations can be improved further through our future explorations.